\pdfoutput=1

\documentclass[11pt]{article}

\usepackage{emnlp2021}

\usepackage{times}
\usepackage{latexsym}
\usepackage{graphicx}
\usepackage{microtype}
\usepackage{amsmath}
\usepackage{booktabs}
\usepackage{dsfont}
\usepackage{amssymb}
\usepackage{subcaption}

\usepackage[T1]{fontenc}
\usepackage[utf8]{inputenc}
\usepackage{microtype}

\title{Disentangling Representations of Text by Masking Transformers}

\author{Xiongyi Zhang \\ Northeastern University \\
\texttt{zhang.xio@northeastern.edu}
         \AND
         Jan-Willem van de Meent \\ University of Amsterdam\\Northeastern University \\  \texttt{j.vandemeent@northeastern.edu} \And
 Byron C. Wallace \\ Northeastern University \\\texttt{b.wallace@northeastern.edu}
 }
\begin{document}
\maketitle
\begin{abstract}
Representations from large pretrained models such as BERT encode a range of features into monolithic vectors, affording strong predictive accuracy across a range of downstream tasks. 
In this paper we explore whether it is possible to learn \emph{disentangled representations} by identifying existing subnetworks within pretrained models that encode distinct, complementary aspects. 
Concretely, we learn binary masks over transformer weights or hidden units to uncover subsets of features that correlate with a specific factor of variation; this eliminates the need to train a disentangled model from scratch for a particular task. 
We evaluate this method with respect to its ability to disentangle representations of \emph{sentiment} from \emph{genre} in movie reviews, \emph{toxicity} from \emph{dialect} in Tweets, and \emph{syntax} from \emph{semantics}.
By combining masking with magnitude pruning we find that we can identify sparse subnetworks within BERT that strongly encode particular aspects (e.g., semantics) while only weakly encoding others (e.g., syntax). 
Moreover, despite \emph{only} learning masks, disentanglement-via-masking performs as well as --- and often better than --- previously proposed methods based on variational autoencoders and adversarial training.
\end{abstract}

\section{Introduction and Motivation}
\label{section:intro}
Large pretrained models such as ELMo \citep{peters2018deep}, BERT \citep{devlin2019bert}, and XLNet \citep{yang2019xlnet} have come to dominate modern NLP. 
Such models rely on self-supervision over large datasets to learn general-purpose representations of text that achieve strong predictive performance across a spectrum of downstream tasks \citep{liu2019roberta}. 
A downside of such learned representations is that it is not obvious what information they encode, which hinders model robustness and interpretability. 
The opacity of embeddings produced by models such as BERT has motivated NLP research on designing probing tasks as a means of uncovering the properties of input texts that are encoded in learned representations \citep{rogers2020primer,linzen2019proceedings,tenney2019bert}.

In this paper we investigate whether we can uncover \emph{disentangled representations} from pretrained models. 
That is, rather than mapping inputs onto a single vector that captures arbitrary combinations of features, our aim is to extract a representation that factorizes into distinct, complementary properties of inputs. 
Explicitly factorizing representations aids interpretability, in the sense that it becomes more straightforward to determine which factors of variation inform predictions in downstream tasks. 

\begin{figure}[!t]
\begin{center}
\includegraphics[width=0.45\textwidth]{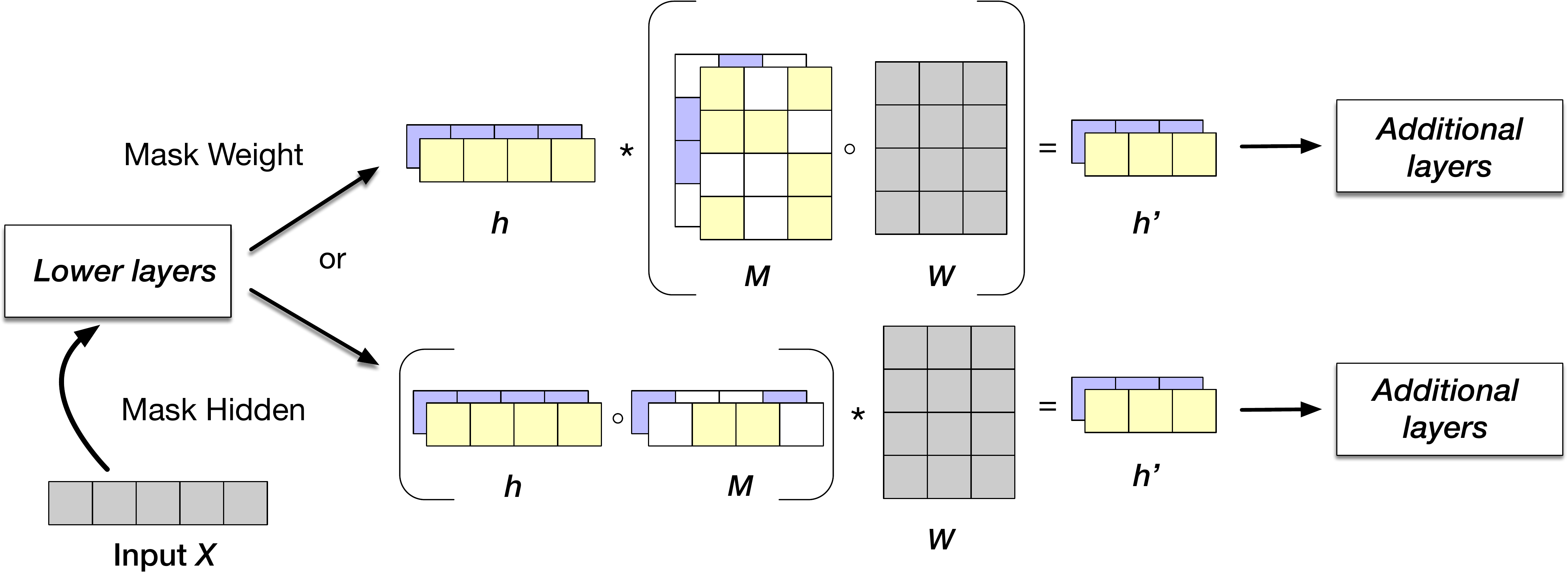}
\end{center}
\caption{Masking weights and hidden activations in BERT. We show a linear layer with weights $W$, inputs $h$, and outputs $h'$. We learn a mask for each disentangled factor, which is either applied to the weights $W$ or to intermediate representations $h$.}
\label{fig:schematic}
\end{figure}

A general motivation for learning disentangled representations is to try and minimize --- or at least expose --- model reliance on spurious correlations, i.e., relationships between (potentially sensitive) attributes and labels that exist in the training data but which are not causally linked \citep{Kaushik2020Learning}. 
This is particularly important for large pretrained models like BERT, as we do not know what the representations produced by such models encode. Here, learning disentangled representations may facilitate increased robustness under distributional shifts by capturing a notion of invariance: If syntactic changes do not affect the representation of semantic features (and vice versa) then we can hope to learn models that are less sensitive to any incidental correlations between these factors.

As one example that we explore in this paper, consider the task of identifying Tweets that contain hate speech \cite{founta2018large}.
Recent work shows that models trained over Tweets annotated on a toxicity scale exhibit a racial bias: They have a tendency to over-predict that Tweets written by users who self-identify as Black are ``toxic'', owing to the use of African American Vernacular English (AAVE; \citealt{sap2019risk}).
In principal, disentangled representations would allow us to isolate relevant signal from irrelevant or spurious factors (such as, in this case, the particular English dialect used), which might in turn reveal and allow us to mitigate unwanted system biases, and increase robustness.

To date, most research on disentangled representations has focused on applications in computer vision \citep{locatello2019challenging,kulkarni2015deep, chen2016infogan,higgins2016beta}, where there exist comparatively clear independent factors of variation such as size, position, and orientation, which have physical grounding and can be formalized in terms of actions of symmetry subgroups \citep{higgins2018definition}. 
A challenge in learning disentangled representations of text is that it is less clear which factors of variation should admit invariance. 
Still, we may hope to disentangle particular properties for certain applications --- e.g., protected demographic information \citep{elazar2018adversarial} --- and there are general properties of language that we might hope to disentangle, e.g., syntax and semantics \citep{chen2019multi}.

\section{Methods}
\label{section:methods}

We are interested in learning a disentangled representation that maps inputs $x$ (text) onto vectors $z^{(a)}$ and $z^{(b)}$ that encode two distinct factors of variation. To do so, we will learn two sets of masks $M^{(a)}$ and $M^{(b)}$ that can be applied to either the weights or the intermediate representations in a pretrained model (in our case, BERT).
We estimate only the mask parameters and do not finetune the weights of the pretrained model.

To learn $M^{(a)}$ and $M^{(b)}$, we assume access to triplets $(x_0, x_1, x_2)$ in which $x_0$ and $x_1$ are similar with respect to aspect $a$ but dissimilar with respect to aspect $b$, whereas $x_0$ and $x_2$ are similar with respect to aspect $b$ but dissimilar with respect to aspect $a$. 
In some of our experiments (e.g., when disentangling sentiment from genre in movie reviews) we further assume that we have access to class labels $y^{(a)}\! \in\! \{0,1\}$ and $y^{(b)}\! \in\! \{0,1\}$ for aspects of interest. In such cases, we build triplets using these labels, defining $(x_0, x_1, x_2)$ such that $y_0^{(a)}=y_1^{(a)} \neq y_2^{(a)}$ and $y_0^{(b)}=y_2^{(b)} \neq y_1^{(b)}$.

\subsection{Masking Weights and Hidden Activations}
Figure \ref{fig:schematic} illustrates the two forms of masking that we consider in our approach (we depict only a single linear layer of the model). 
Here $h = (h^{(a)}, h^{(b)})$ are input activations, $W$ are the weights in the pretrained model,\footnote{We omit the bias term, which we do not mask.} and $h'=(h'^{(a)}, h'^{(b)})$ are output activations. 
We augment each layer of the original network with two (binary) masks $M=(M^{(a)}, M^{(b)})$, applied in one of two ways:

\paragraph{1. Masking Weights} Here masks $M^{(a)}$ and $M^{(b)}$ have the same shape as weights $W$, and outputs are computed using the masked weights tensor
\begin{align}
    h' = h \cdot (W \circ M).
\end{align}

\paragraph{2. Masking Hidden Activations} In this case masks $M^{(a)}$ and $M^{(b)}$ have the same shape as the intermediate (hidden) activations $h^{(a)}$ and $h^{(b)}$. Output activations are computed by applying the original weights $W$ to masked inputs
\begin{align}
    h' = (h \circ M) \cdot W.
\end{align}

In both methods, we follow \citep{zhao2020masking} and only mask the last several layers of BERT, leaving bottom layers 
unchanged.\footnote{We mask the last six layers, which seems to work well.}

\subsection{Triplet Loss}
\label{section:triplet-loss}

To learn masks, we assume that we have access to supervision in the form of triplets, as introduced above. 
Passing $(x_{0}, x_{1}, x_{2})$ through our model yields two representations for each instance: $(z^{(a)}_{0},z^{(b)}_{0}) ,(z^{(a)}_{1},z^{(b)}_{1}), (z^{(a)}_{2},z^{(b)}_{2})$, for which we define the losses
\begin{align}
    \nonumber
    \mathcal{L}^{(a)}_{\text{trp}} 
    &= 
    \max
    \Big(
      \|z^{(a)}_{0} \!\!\!-\! z^{(a)}_{1}\| - \|z^{(a)}_{0} \!\!\!- \! z^{(a)}_{2}\| + \alpha, 0
    \Big), \\
    \nonumber
    \mathcal{L}^{(b)}_{\text{trp}} 
    &= 
    \max
    \Big(
        \|z^{(b)}_{0} \!\!\!-\! z^{(b)}_{2}\| - \|z^{(b)}_{0} \!\!\!-\! z^{(b)}_{1}\| + \alpha, 0
    \Big), \\
    \mathcal{L}_{\text{trp}} 
    &= 
    \frac{1}{2} 
    \Big(
        \mathcal{L}^{(a)}_{\text{trp}} + \mathcal{L}^{(b)}_{\text{trp}}
    \Big).
\end{align}
Here $\alpha$ is a hyperparameter specifying a margin for the loss, which we set to $\alpha=2$ in all experiments.

\subsection{Supervised Loss}
\label{section:additional-losses}

In some settings we may have access to more direct forms of supervision. For example, when learning representations for the genre and sentiment in a movie review, we have explicit class labels $y^{(a)}$ and $y^{(b)}$ for each aspect. 
To exploit such supervision when available, we add classification layers $C^{(a)}$ and $C^{(b)}$ and define classification losses 
\begin{gather}
    \mathcal{L}^{(a)}_{\text{cls}} 
    = 
    \text{CrossEntropy}
    \Big(
      C^{(a)}(z^{(a)}), y^{(a)}
    \Big),  \\
    \mathcal{L}^{(b)}_{\text{cls}} 
    = \text{CrossEntropy}
    \Big(
      C^{(b)}(z^{(b)}), y^{(b)}
    \Big), \\
    \mathcal{L}_{\text{cls}} 
     = 
    \frac{1}{2} 
    \Big(
      \mathcal{L}^{(a)}_{\text{cls}} + \mathcal{L}^{(b)}_{\text{cls}}
     \Big).
\end{gather}

\subsection{Disentanglement Loss}
\label{section:disentanglement-loss}

To ensure that the two aspect representations are distinct, we encourage the masks to 
overlap as little as possible. 
To achieve this we add a term in the loss for each layer $l \in L$ 
\begin{equation}
    \mathcal{L}_{\text{ovl}} = \frac{1}{|L|}\sum_{l\in L}\sum_{i,j}\mathds{1}_{(M^{(a)}_{i,j} + M^{(b)}_{i,j} > 1)}
    .
\end{equation}

\subsection{Binarization and Gradient Estimation}

The final loss of our model is
\begin{equation} \label{eq:loss}
    \mathcal{L} = \lambda_{\text{trp}} \cdot \mathcal{L}_{\text{trp}} + \lambda_{\text{ovl}} \cdot \mathcal{L}_{\text{ovl}}\:\: (+ \lambda_{\text{cls}} \cdot \mathcal{L}_{\text{cls}}).
\end{equation}
\noindent We parenthetically denote the classification loss, which we only include when labels are available. We minimize this loss to estimate $M$ (and classifier parameters), keeping the pretrained BERT weights fixed. Because the loss is not differentiable with respect to a binary mask, we learn continuous masks $M$ that are binarized during the forward pass by applying a threshold $\tau$, a global hyperparameter,
\begin{equation} 
    M^*_{ij} = 
    \begin{cases}
      1 & \text{if } M_{ij} \geq \tau \\
      0 & \text{if } M_{ij} < \tau\\
    \end{cases}     
    .
\end{equation}
We then use a \emph{straight-through} estimator \citep{hinton2012neural,bengio2013estimating} to approximate the derivative, which is to say that we evaluate the derivative of the loss with respect to the continuous mask $M$ at the binarized values $M=M^*$,
\begin{equation}
    M = M - \eta \left. \frac{\partial L}{\partial M} \right\vert_{M=M^{*}}.
\end{equation} 
\section{Experiments}
\label{section:experiments}
We conduct a series of experiments to evaluate the degree to which the proposed masking strategy achieves disentanglement, as compared to existing methods for disentanglement in NLP.
As a first illustrative example, we consider a corpus of movie reviews, in which sentiment is correlated with film genre (\ref{section:movies}). 
We treat this as a proxy for a spurious correlation, and evaluate the robustness of the models to shifts in conditional probabilities of one attribute (sentiment) given another (genre).
We then consider a more consequential example: Hate speech classification on Twitter (\ref{section:toxic}). 
Prior work \cite{sap2019risk} has shown that models exploit a spurious correlation between ``toxicity'' and African American Vernacular English (AAVE); we aim to explicitly disentangle these factors in service of fairness. We evaluate whether the model is able to achieve \emph{equalized odds}, a commonly used fairness metric.
Finally, following prior work, we investigate disentangling semantics from syntax (insofar as this is possible) in Section \ref{section:semsyn}.

\subsection{Disentangling Sentiment From Genre}
\label{section:movies}

\begin{table}
\centering
\small
    \begin{center}
    \textbf{IMDB}\\[8pt]
    \end{center}
    \begin{tabular}{ llrr } 
      \toprule
      & Sentiment /\ Genre & Drama & Horror \\ 
      \midrule
      Original &Positive & 41.2 & 7.4 \\ 
        Dataset &Negative & 20.0 & 31.4 \\ 
      \midrule
      Our Train &Positive & 42.5 & 7.5 \\ 
      (Correlated) &Negative & 7.5 & 42.5 \\ 
    \midrule
      Our Test &Positive & 25.0 & 25.0 \\ 
      (Uncorrelated) & Negative & 25.0 & 25.0 \\
      \bottomrule
    \end{tabular}\\
    \begin{center}
    \textbf{Twitter}\\[8pt]
    \end{center}
    \begin{tabular}{ llrr } 
      \toprule
      & Toxicity /\ Race\hspace{12 pt} & Black & White \\ 
      \midrule
      Original &Toxic\hspace{24 pt} & 15.4 & 16.9 \\ 
        Dataset &Non-toxic & 20.8 & 46.9 \\ 
      \midrule
      Train &Toxic\hspace{24 pt} & 42.5 & 7.5 \\ 
      (Correlated) &Non-toxic & 7.5 & 42.5 \\ 
    \midrule
      Test &Toxic\hspace{24 pt} & 25.0 & 25.0 \\ 
      (Uncorrelated) & Non-toxic & 25.0 & 25.0 \\ 
      \bottomrule
    \end{tabular}\\
    \caption{Percentage of each class in the original dataset, and in the two subsets we sampled, one correlated training set and one uncorrelated test set. We train models on the former and test on the latter. This is meant to assess the robustness of models to shifts in spurious correlations that might exist in training data.}
     \label{table:setup}
\end{table}

\paragraph{Experimental Setup} In this experiment we assume a setting in which each data point $x$ has both a `main' label $y$ and a secondary (possibly sensitive) attribute $z$. 
We are interested in evaluating the degree to which explicitly disentangling representations corresponding to these may afford robustness to shifts in the conditional distribution of $y$ given $z$. 
As a convenient, illustrative dataset with which to investigate this, we use a set of movie reviews from IMDB \citep{maas2011learning} in which each review has both a binary sentiment label and a genre label.

We pick the two genres of movies that exhibit a strong correlation with review sentiment: Drama (reviews tend to be positive) and Horror (negative), excluding reviews corresponding to other genres and the (small) set of instances that belong to both genres. 
To investigate robustness to shifts in correlations between $z$ and $y$ we sampled two subsets from the training set such that in the first sentiment and genre are highly correlated, while in the second they are uncorrelated. 
We report the correlations between these variables in the two subsets in Table \ref{table:setup}. 
We train models on the correlated subset, and then evaluate them on the uncorrelated set. 

\begin{figure}
\centering
\includegraphics[width=0.5\textwidth]{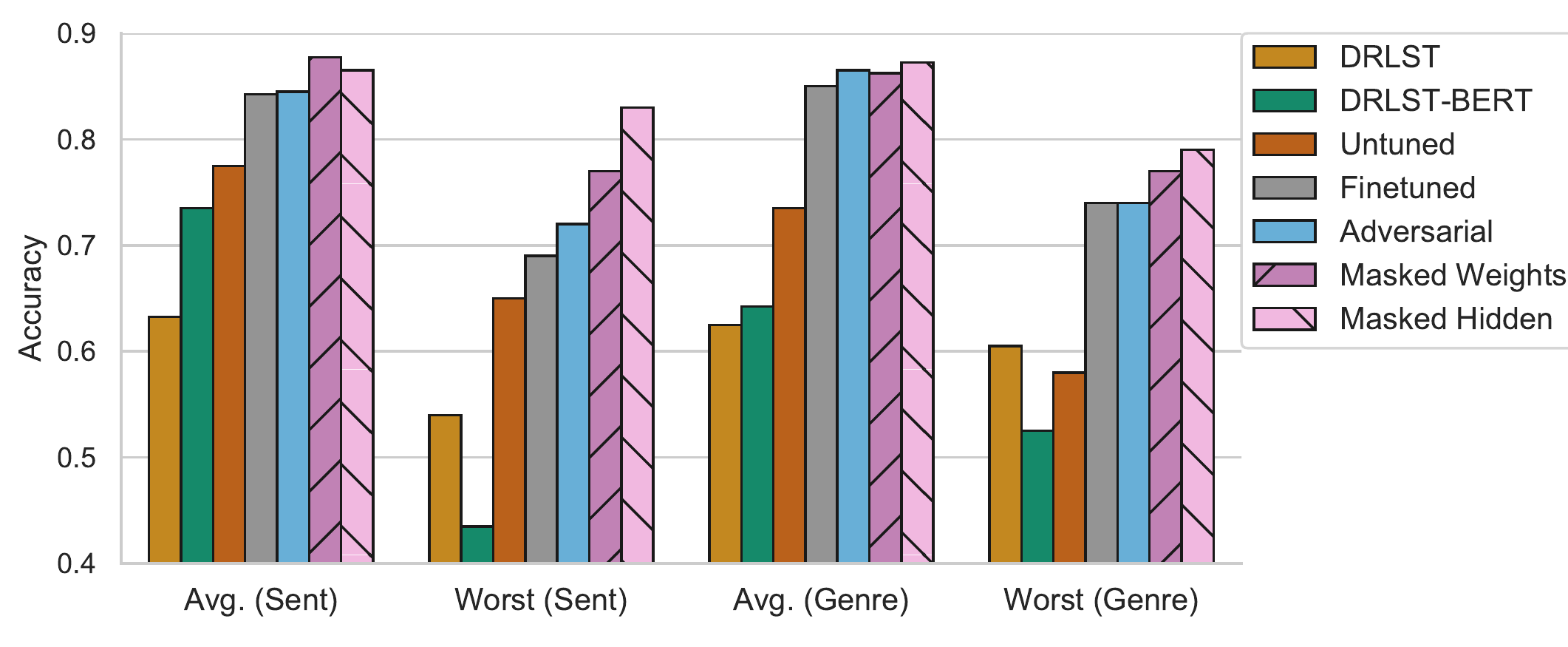}
\caption{Average and worst main task performance across sentiment/genre combinations. Masked variants (proposed in this paper) are cross-hatched. Large gaps between average and worst performance for a model suggest that it is using the non-target attribute when making predictions for the main task.}
\label{fig:imdb_avg_worst}
\end{figure}

\definecolor{plotBlue}{RGB}{0,114,180}
\definecolor{plotYellow}{RGB}{230,160,0}

\begin{figure*}
\begin{center}
\includegraphics[width=.775\textwidth]{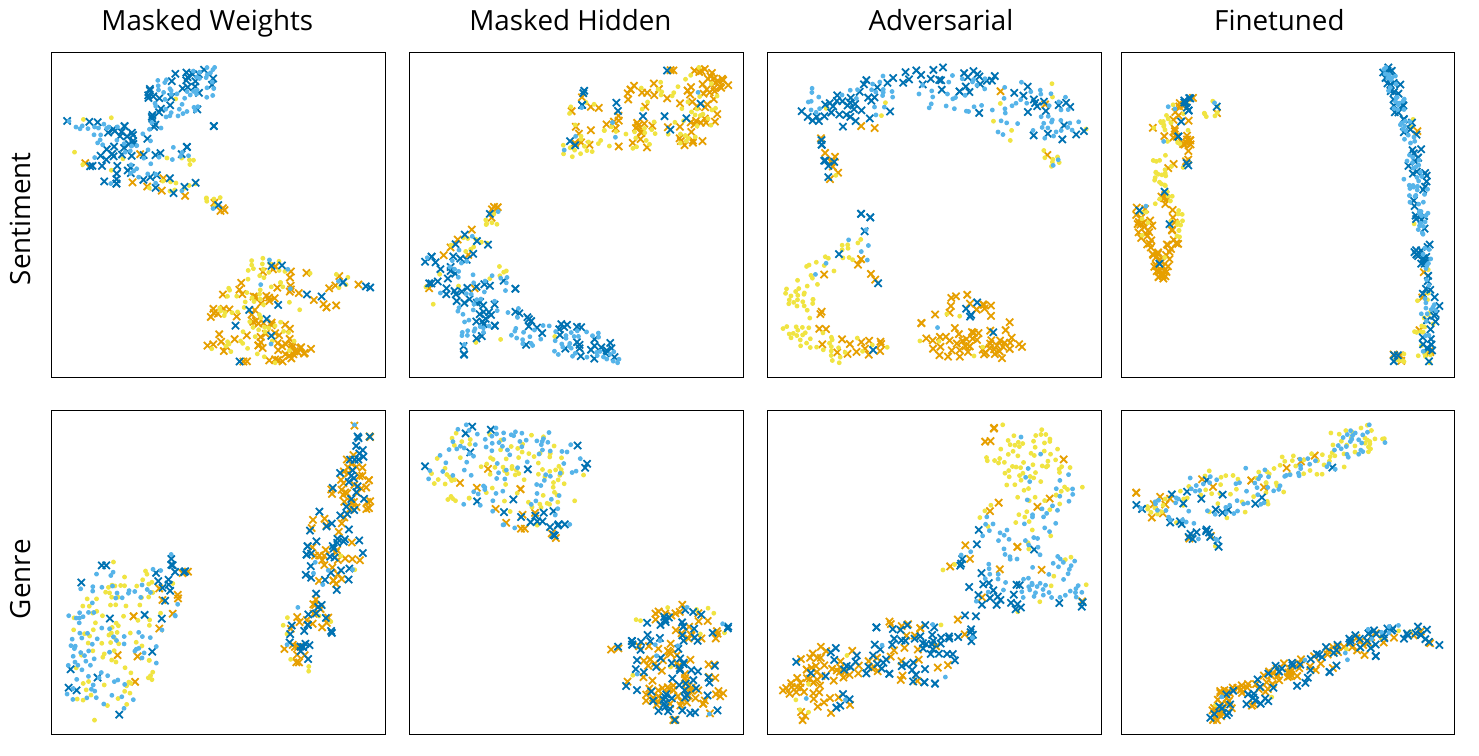}
\end{center}
\caption{t-SNE projection of sentiment and genre representations of different models. Marker colors denote sentiment (\textbf{\color{plotBlue} blue} for positive and \textbf{\color{plotYellow} yellow} for negative); marker shapes denote genre ($\times$ for drama and $\bullet$ for horror). In the upper row we expect the points of the same color to be clustered together, but not the points with the same marker shapes, and for the lower row we expect the points of the same  marker shapes to be clustered together, but not those of the same colors.}
\label{fig:tsne}
\end{figure*}

We compare the proposed masking approaches to several baselines. 
\textbf{Untuned} is a dense classification layer on top of BERT representations (without finetuning). 
In the \textbf{finetuned} variant we omit masks and instead minimize the loss with respect to BERT weights. 
In the \textbf{adversarial} model we adopt `adversarial debiasing': In addition to minimizing loss on the main task, we train an adversarial classifier to predict the non-target attribute, and the encoder is trained to mitigate the adversaries' ability to do so. 
We implement this via gradient-reversal \cite{ganin2015unsupervised}. 
We also compare to two variational autoencoder baselines: \textbf{DRLST} \cite{john2019disentangled} is a VAE model with multi-task loss and adversarial loss; and \textbf{DRLST-BERT} is the same model, except we use BERT as the encoder in place of a GRU \cite{cho2014properties}.

\begin{table}
\small
\centering
       \begin{tabular}{ lcc } 
       \toprule

         & \textbf{Sentiment} $\uparrow$ & \textbf{Leakage (Genre)} $\downarrow$  \\
        \midrule
     DRLST & 62.1 &59.0  \\ 
     DRLST-BERT & 67.5 & 66.3 \\ 
        \midrule
     Untuned & 82.3 & 81.5 \\ 
     Finetuned & 87.5 & 85.5   \\ 
     Adversarial  & 86.8 & 80.3    \\ 
        \midrule
     Masked Weights& 88.0 & 72.0  \\ 
     Masked Hidden & 88.0  & 79.0 \\	
        \bottomrule
    \end{tabular}
\caption{Performance on the main task of sentiment analysis and gender information leakage. The DRLST baselines perform poorly on the main task; the proposed masking approaches have achieve comparable results for sentiment, and expose less genre information.  }
    \label{table:leak-imdb-sent}
\end{table}

\paragraph{Leakage of the Non-target Attribute} We evaluate the degree to which representations ``leak'' non-target information. 
Following \cite{elazar2018adversarial}, we first train the model to predict the main task label on the correlated dataset. 
Then we fix the encoder and train a single layer MLP on the uncorrelated dataset to probe the learned representations for the non-target attribute. 
Because this probe is trained and tested on only uncorrelated data, it cannot simply learn the main task and exploit the correlation.
We report results for our proposed masking models and baselines in Table \ref{table:leak-imdb-sent}. We also report the results with genre classification as the main task and sentiment as the protected attributes in the Appendix (Section \ref{sec:imdb-genre}).
The DRLST baselines generally underperform, which translates to low leakage numbers but also poor performance on the main task.
Compared to the baselines, our masking variants perform comparably with respect to predicting the main task label, but do so with less leakage. 

\paragraph{Worst Group Performance} In addition to non-target attribute leakage, we measure how models perform on the main task for each subgroup: (Positive, Drama), (Positive, Horror), (Negative, Drama), and (Negative, Horror).
Because the distribution of the four groups is unequal in the train set, we expect that models will perform better on attribute combinations that are over-represented in this set, and worse on those that are under-represented, suggesting that the model is implicitly exploiting the correlation between these attributes. 
We report both the average and worst performance on the four subgroups; the latter is a proxy to measure robustness when subgroup compositions shift between the train to the test set.

Figure \ref{fig:imdb_avg_worst} plots the results.
We observe that the masking variants realize similar average performance as the baselines, but consistently outperform these in terms of worst performance. 
This indicates that the proposed variants rely less on the correlation between the two attributes when predicting the main label.

\definecolor{plotBlue}{RGB}{0,114,180}
\definecolor{plotYellow}{RGB}{230,160,0}

\paragraph{Qualitative Evaluation} In Figure \ref{fig:tsne} we plot t-SNE visualizations \citep{maaten2008visualizing} of the representations induced by different models. 
If the representations are disentangled as desired, instances with different sentiment will be well separated, while those belonging to different genres within each sentiment will \textbf{not} be separated. 

Similarly, for genre representations, instances of the same genre should co-locate, but clusters should not reflect sentiment. 
No method perfectly realizes these criteria, but the proposed masking approaches achieve better results than do the two baselines. 
For instance, in the embeddings from the adversarial (Sentiment) and finetuned (Sentiment), instances that have negative sentiment but different genres (\textbf{\color{plotYellow} $\bullet$} and \textbf{\color{plotYellow} $\times$}) are separated, indicating that these sentiment representations still carry genre information.

\subsection{Disentangling Toxicity from Dialect}
\label{section:toxic} 

\paragraph{Experimental Setup} In this experiment we evaluate models on a more consequential task: Detecting hate speech in Tweets \cite{founta2018large}. 
Prior work \cite{sap2019risk} has shown that existing hate speech datasets exhibit a correlation between African American Vernacular English (AAVE) and toxicity ratings, and that models trained on such datasets propagate these biases. 
This results in Tweets by Black individuals being more likely to be predicted as ``toxic''. 
Factorizing representations of Tweets into dialectic and toxicity subvectors could ameliorate this problem.

We use \cite{founta2018large} as a dataset for this task. 
This comprises 100k Tweets, each with a label indicating whether the Tweet is considered toxic, and self-reported information about the author. 
We focus on the self-reported race information.
Specifically, we subset the data to include only users who self-reported as being either white or Black.
The idea is that Tweets from Black individuals will sometimes use AAVE, which in turn could be spuriously associated with `toxicity'.

Similar to the above experiment, we sampled two subsets of the data such that in the first the (annotated) toxicity and self-reported race are highly correlated, while in the second they are uncorrelated 
(see Table \ref{table:setup}).
We train models on the correlated subset, and evaluate them on the uncorrelated set. 
This setup is intended to measure the extent to which models are prone to exploiting (spurious) correlations, and whether and which disentanglement methods render models robust to these.

\paragraph{Leakage of Race Information} We evaluate the degree to which representations of Tweets ``leak'' information about the (self-reported) race of their authors using the same method as above, and report results in Table \ref{table:leak-tox}.
We observe that the proposed masking variants perform comparably to baselines with respect to predicting the toxicity label, but leak considerably less information pertaining to the sensitive attribute (race). 
\paragraph{Fairness Implications}
In addition to the degree to which representations encode race information, we are interested in how the model performs on instances comprising (self-identified) Black and white individuals, respectively. 
More specifically, we can measure the \textbf{True Positive Rate (TPR)} and the \textbf{True Negative Rate (TNR)} on these subgroups, which in turn inform \emph{equalized odds}, a standard metric used in the fairness literature.

\begin{table}
\small
\centering

       \begin{tabular}{ lcc } 
       \toprule

         & \textbf{Toxicity} $\uparrow$ & \textbf{Leakage (Race)} $\downarrow$  \\
        \midrule
     DRLST & 66.4 &86.6 \\ 
     DRLST-BERT & 68.0 &90.0 \\ 
        \midrule
     Untuned  & 68.2 & 76.8  \\ 
     Finetuned & 70.4 & 93.0   \\ 
     Adversarial & 70.2 & 67.0   \\ 
        \midrule
     Masked Weights & 70.4 & \textbf{59.8}  \\ 
     Masked Hidden &\textbf{71.4}  & 63.5 \\	
        \bottomrule
    \end{tabular}
\caption{Performance on the main task of toxicity prediction, and leakage of race information. Compared to the baselines, the proposed approaches achieve performance competitive with or better than baselines, while minimizing leakage of the protected attribute.}
    \label{table:leak-tox}
\end{table}

\begin{figure*}
\centering
\includegraphics[width=0.8\textwidth]{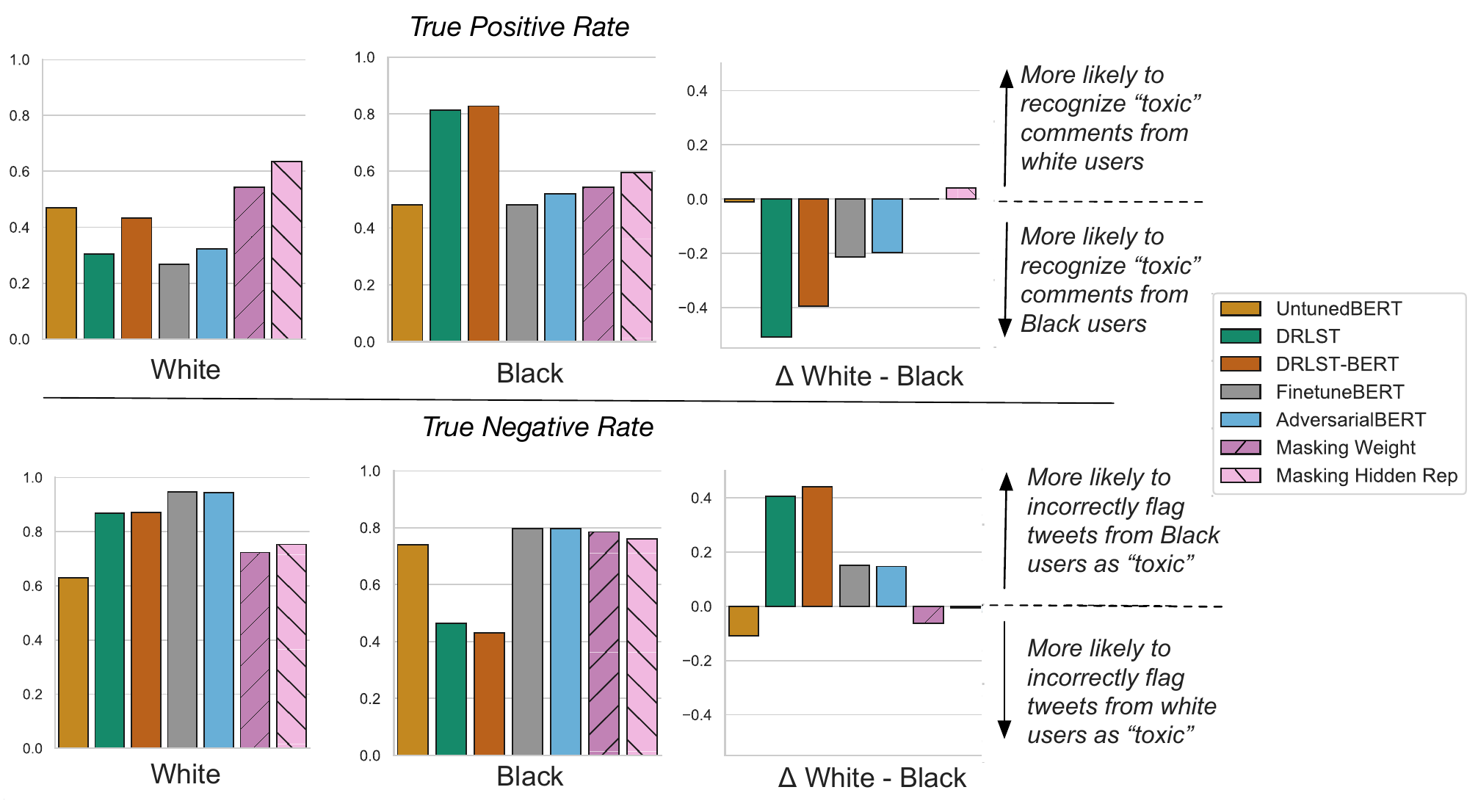}
\caption{True Positive and True Negative Rates achieved on white and Black individuals, respectively, and the (signed) difference between these (rightmost subplots). The proposed masked variants (cross-hatched) are more equitable in performance, while other methods tend to over-predict Tweets written by Black individuals as ``toxic".}
\label{fig:tox_group}
\end{figure*}

We report the TPR and TNR of each model achieved over white and Black individuals, respectively, as well as the difference across the two groups in Figure \ref{fig:tox_group}. 
We observe that the proposed model variants achieve a smaller TPR and TNR gap across the two races (see rightmost subplots), indicating that performance is more equitable across the groups, compared to baselines.

\subsection{Disentangling Semantics from Syntax}
\label{section:semsyn}

\paragraph{Experimental Setup} As a final experiment, we follow prior work in attempting to disentangle semantic from syntactic information encoded in learned (BERT) representations of text.
Because we have proposed exploiting triplet-loss, we first construct triplets $(x_{0}, x_{1}, x_{2})$ such that $x_{0}$ and $x_{1}$ are similar semantically but differ in syntax, while $x_{0}$ and $x_{2}$ are syntactically similar but encode different semantic information. 
We follow prior work \citep{chen2019multi,ravfogel2019unsupervised} in deriving these triplets.
Specifically, we obtain $x_{0}, x_{1}$ from the  ParaNMT-50M \citep{wieting-gimpel-2018-paranmt} dataset. 
Here $x_{1}$ is obtained by applying back-translation to $x_{0}$, i.e., by translating $x_{0}$ from English to Czech and then back into English. 
To derive $x_{2}$ we keep all function words (from a list introduced in \citealt{ravfogel2019unsupervised})
in $x_{0}$, and replace content words by masking each in turn, running the resultant input forward through BERT, and randomly selecting one of the top predictions (that differs from the original word) as a replacement.

We compare our disentanglement-via-masking strategies against models that represent state-of-the-art approaches to disentangling syntax and semantics.
In particular, we compare against VGVAE \citep{chen2019multi}, though we implement this on top of BERT-base to allow fair comparison.
Following prior work that has used triplet loss for disentanglement, we also compare against a model in which we finetune BERT using the same triplet loss that we use to train our model, but in which we update all model parameters (as opposed to only estimating mask parameters).
To evaluate learned representations with respect to the semantic and syntactic information that they encode, we evaluate them on four tasks. 
Two of these depend predominantly on semantic information, while the other two depend more heavily on syntax.\footnote{This is a (very) simplified view of `semantics' / `syntax'.}
For the semantics tasks we use: (i) A word content (WC) \citep{conneau-etal-2018-cram} task in which we probe sentence representations to assess whether the corresponding sentence contains a particular word; and (ii) A semantic textual similarity (STS) benchmark \citep{rosenthal2017semeval}, which includes human provided similarity scores between pairs of sentences. 
We evaluate the former in terms of accuracy; for the latter (a ranking task) we use Spearman correlation.
To evaluate whether representations encode syntax, we use: (i) A task in which the aim is to predict the length of the longest path in a sentence's parse tree from its embedding (Depth) \citep{conneau-etal-2018-cram}; and (ii) A task in which we probe sentence representations for the type of their top constituents immediately below the $S$ node (TopConst).\footnote{See \cite{conneau-etal-2018-cram} for more details regarding WC, Depth, and TopConst tasks.}

\begin{figure}
\begin{center}
\includegraphics[width=.4\textwidth]{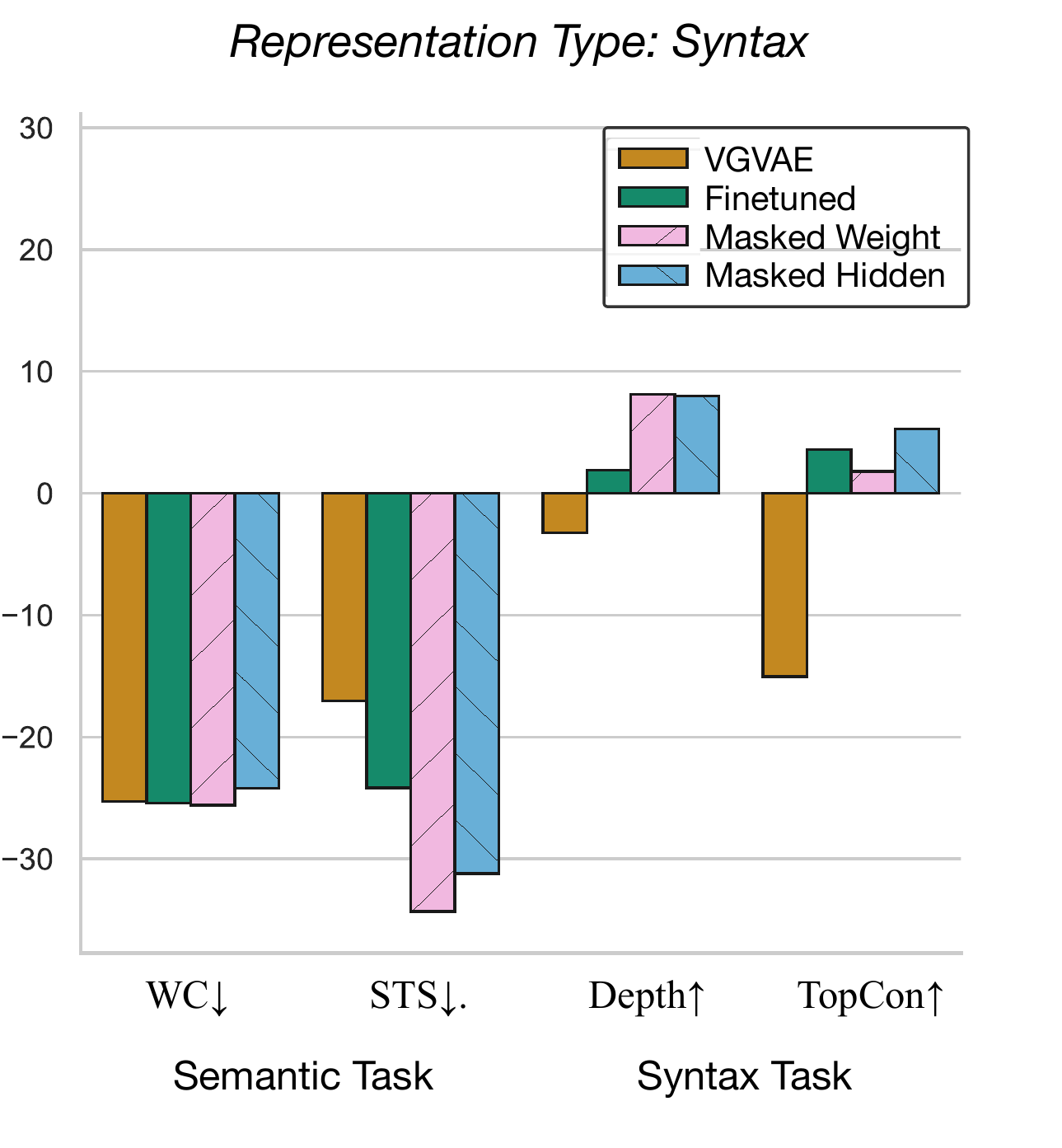}
\end{center} 
\caption{Differences between performances achieved via BERT embeddings and the disentangled model variants considered on semantics-oriented (WC, STS) and syntax-oriented (Depth, TopCon) tasks compared with BERT embeddings. We plot this difference with respect to the syntax embeddings induced by the models.}
\label{fig:rep-syn}
\end{figure}

Figure \ref{fig:rep-syn} 
shows the signed differences between the performance achieved on semantics- and syntax-oriented tasks by BERT embeddings (we mean-pool over token embeddings) and the `syntax' representations from the disentangled models considered (see the Appendix for the analogous plot for the `semantics' representations in figure \ref{fig:rep-sem}).
Ideally, syntax embeddings would do well on the syntax-oriented tasks (Depth and TopCon) and poorly on the semantic tasks (WC and STS). 
With respect to syntax-oriented tasks, the proposed masking methods outperform BERT base representations, as well as the alternative disentangled models considered.
These methods also considerably reduce performance on semantics-oriented tasks, as we would hope.

We emphasize that this is achieved \emph{only via masking, and without modifying the underlying model weights}.

\subsection{Identifying \emph{Sparse} Disentangled Sub-networks for Semantic and Syntax}

We next assess if we are able to identify \emph{sparse} disentangled subnetworks by combining the proposed masking approaches with \emph{magnitude pruning} \citep{han2015deep}.
Specifically, we use the loss function defined in Equation \ref{eq:loss} to finetune BERT for $k$ iterations, and prune weights associated with the $m$ smallest magnitudes after training. 
We then initialize masks to the sparse sub-networks identified in this way, and continue refining these masks via the training procedure proposed above.
We compare the resultant sparse network to networks similarly pruned (but not masked). 
Specifically, for the latter we consider: Standard magnitude tuning applied to BERT, without additional tuning (Pruned + Untuned), and a method in which after magnitude pruning we resume finetuning of the subnetwork until convergence, using the aforementioned loss function (Pruned + Finetuned).

We compare the performance achieved on the semantic and syntax tasks by the subnetworks identified using the above strategies at varying levels of sparsity, namely after pruning: \{0, 20\%, 40\%, 60\%, 80\%, 85\%, 90\%, 95\%\} of weights.\footnote{Technically, in the Pruned + Masked Weights method, refining the masks may change subnetwork sparsity, but empirically we find this to change the sparsity only slightly ($\sim$1\% in all of our experiments).}
We report full results in Appendix Figure \ref{fig:prune}, but here observe that combining the proposed masking strategy with magnitude pruning consistently yields representations of semantics that perform comparatively strongly on the semantics-oriented tasks (STS, WC), even at very high levels of sparsity; these semantics representations also perform comparatively poorly on the syntax-oriented tasks (Depth, TopCon), as one would hope. 
Similarly, syntax representations perform poorly on semantics-oriented tasks, and outperform alternatives on the syntax-oriented tasks. 
In sum, this experiment suggests that we are indeed able to identify \emph{sparse} disentangled subnetworks via masking.

\section{Related Work} 
\label{section:related work}

\paragraph{Disentangled and structured representations of images.} The term \emph{disentangled representations} has been used to refer to a range of methods with differing aims.
Much of the initial focus in this space was on learning representations of images, in which certain dimensions correspond to interpretable factors of variation \citep{kulkarni2015deep,higgins2016beta,chen2016infogan}. In the context of variational autoencoders \citep{kingma2013auto,rezende2014Stochastic} this motivated work that evaluates to what extent such representations can recover a set of ground-truth factors of variation when learned without supervision \citep{eastwood2018evaluation,kim2018disentangling,chen2018isolating}. Other work has investigated representations with the explicit motivation of fairness \citep{locatello2019fairness,creager2019flexibly}, which disentanglement may help to facilitate. 

\paragraph{Disentangling representations in NLP.} Compared to vision, there has been relatively little work on learning disentangled representations of text. 
Much of the prior work on disentanglement for NLP that does exist has focused on using such representations to facilitate \emph{controlled generation}, e.g., manipulating sentiment \citep{larsson2017disentangled}. 

A related notion is that of \emph{style transfer}, for example, separating style from content in language models \citet{shen2017style,mir2019evaluating}.
There has also been prior work on learning representations of particular aspects to facilitate domain adaptation \citep{zhang2017aspect}, and aspect-specific information retrieval \citep{jain2018learning}. \citet{esmaeili2019structured} focus on disentangling user and item representations for product reviews. \citet{moradshahi2019hubert} combine BERT with Tensor-Product Representations to improve its transferability across different tasks.
Recent work has proposed learning distinct vectors coding for semantic and syntactic properties of text \citep{chen2019multi,ravfogel2019unsupervised};
these serve as baseline models in our experiments.

Finally, while not explicitly framed in terms of disentanglement, efforts to `de-bias' representations of text are related to our aims. 
Some of this work has used adversarial training to attempt to remove sensitive information \citep{elazar2018adversarial,barrett2019adversarial}. 

\paragraph{Network pruning.} A final thread of relevant work concerns selective pruning of neural networks. This has often been done in the interest of model compression \cite{han2015deep,han2015learning}.
Recent intriguing work has considered pruning from a different perspective: Identifying small subnetworks --- winning `lottery tickets' \citep{frankle2018lottery} --- that, trained in isolation with the right initialization, can match the performance of the original networks from which they were extracted. Very recent work has demonstrated that winning tickets exist within BERT \citep{chen2020lottery}.

\section{Discussion}
\label{section:discussion}
We have presented a novel perspective on learning disentangled representations for natural language processing in which we attempt to uncover existing subnetworks within pretrained transformers (e.g., BERT) that yield disentangled representations of text.
We operationalized this intuition via a \emph{masking} approach, in which we estimate \emph{only} binary masks over weights or hidden states within BERT, leaving all other parameters unchanged. 
We demonstrated that --- somewhat surprisingly --- we are able to achieve a level of disentanglement that often exceeds existing approaches (e.g., a varational auto-encoder on top of BERT), which have the benefit of finetuning all model parameters.

Our experiments demonstrate the potential benefits of this approach. In Section \ref{section:movies} we showed that disentanglement via masking can yield representations that are comparatively robust to shifts in correlations between (potentially sensitive) attributes and target labels. 
Aside from increasing robustness, finding sparse subnetworks that induce disentangled representations constitutes a new direction to pursue in service of providing at least one type of model interpretability for NLP. 
Finally, we note that sparse masking (which does not mutate the underlying transformer parameters) may offer efficiency advantages over alternative approaches.

\section{Acknowledgements}

This work was supported by that National Science Foundation (NSF), grant 1901117. 

\bibliography{disentangle_anthology,disentangle_custom_new}
\bibliographystyle{acl_natbib}

\appendix
\renewcommand\thetable{\thesection.\arabic{table}}    
\renewcommand{\thefigure}{A.\arabic{figure}}

\clearpage
\section{Appendix}
\label{sec:appendix}
\setcounter{figure}{0}    
\setcounter{table}{0}    

\subsection{Additional IMDB Results}
\label{sec:imdb-genre}

In Table \ref{table:leak-imdb-genre} we report results treating genre classification as the main task, and sentiment as the `protected' attribute.

\begin{table}[hb]
\small
\centering

       \begin{tabular}{ lcc } 
       \toprule

         & \textbf{Genre} $\uparrow$ & \textbf{Leakage (Sentiment)} $\downarrow$  \\
        \midrule
     DRLST & 65.6 & 61.1  \\ 
     DRLST-BERT & 71.4 &70.3 \\ 
        \midrule
     Untuned & 81.5 & 82.3 \\ 
     Finetuned & 87.3 & 86.0   \\ 
     Adversarial  & 85.0 & 75.5    \\ 
        \midrule
     Masked Weights& 87.0 & 73.0  \\ 
     Masked Hidden & 85.0  & 79.0 \\	
        \bottomrule
    \end{tabular}
\caption{Performance of models when treating genre prediction as the main task and sentiment as the `protected' variable. The DRLST baselines perform poorly on the main task. The proposed masking methods achieve performance comparable to other baselines, while encoding less of the sentiment information.}
    \label{table:leak-imdb-genre}
\end{table}

\subsection{Distribution of Learned Masks Across BERT Layers }
\label{sec:mask-mha}

\begin{figure}
    \centering
    \label{fig:mask_dist}
    \begin{subfigure}[ht]{0.4\textwidth}
        \centering
        \includegraphics[height=1.2in]{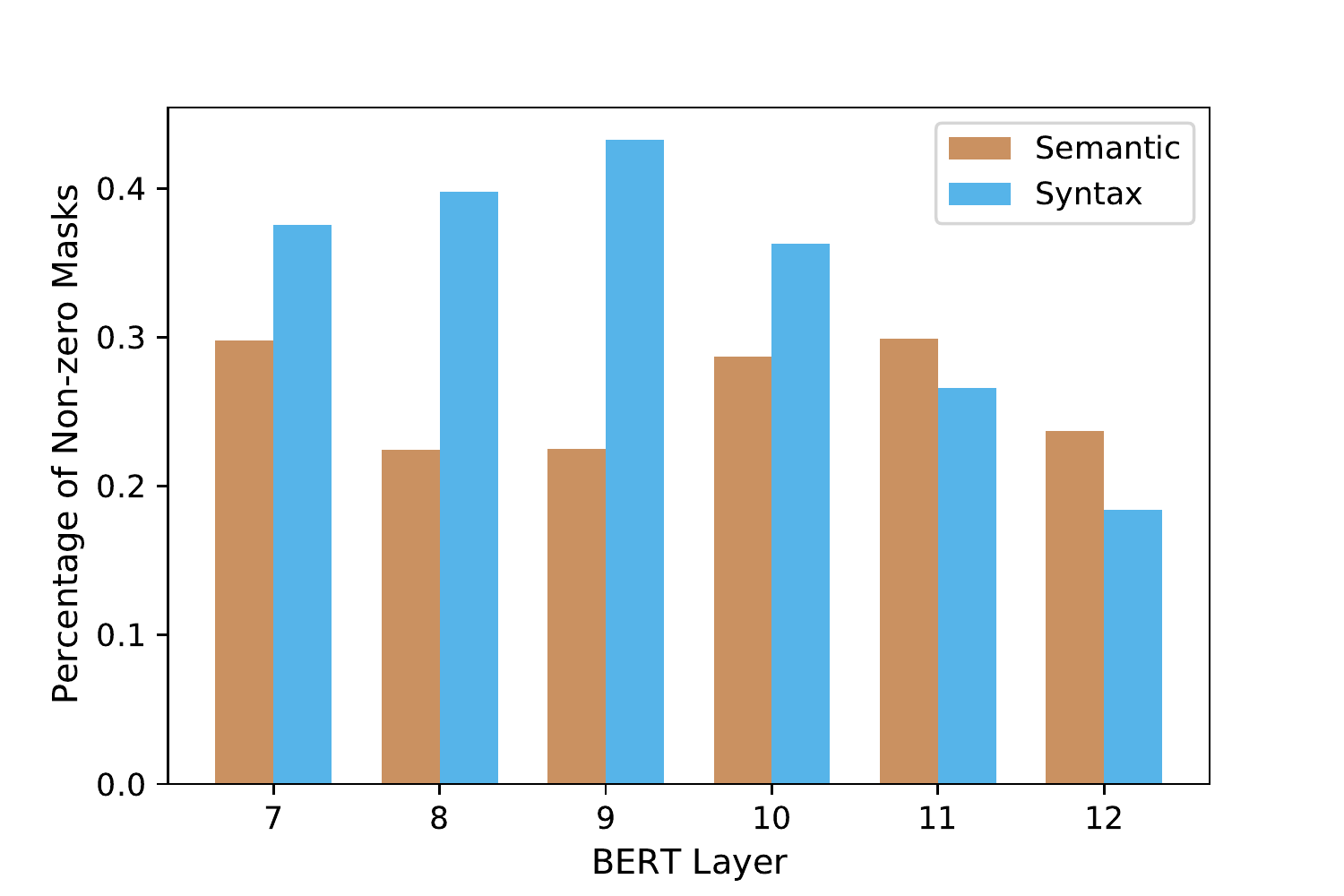}
        \caption{Masking Weights}
    \end{subfigure}%
    \\
    \begin{subfigure}[ht]{0.4\textwidth}
        \centering
        \includegraphics[height=1.2in]{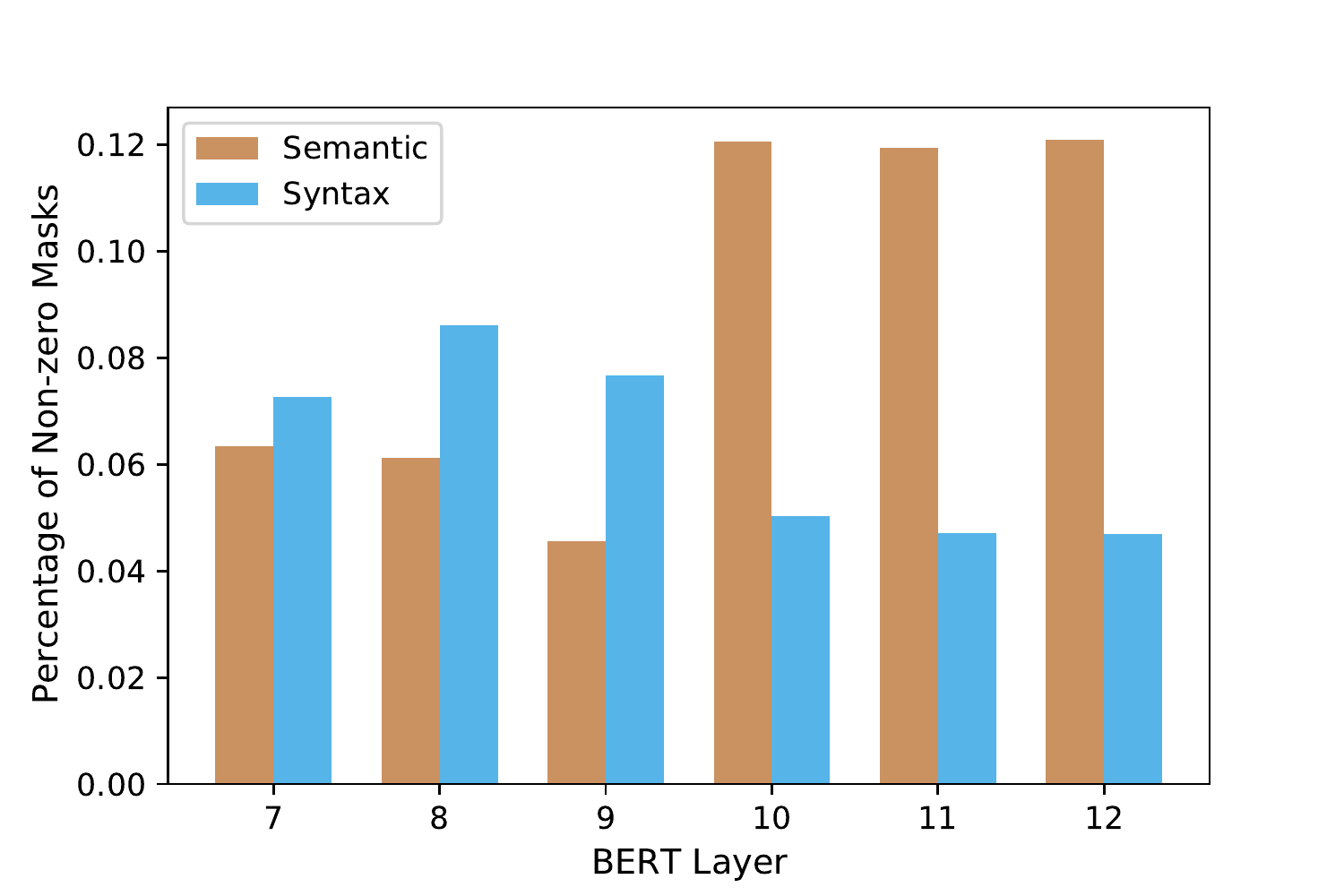}
        \caption{Masking Hidden Activations}
    \end{subfigure}
    \caption{Non-zero (unmasked) elements of the semantic and syntax subnetworks induced by masking weights (a) and hidden representations / activations (b).} 
    \label{fig:sparse}
\end{figure}

Here we inspect the subnetworks (i.e., the weights or hidden activations that are not masked) uncovered by our model, which may provide insights regarding where pretrained (masked) language models encode different sorts of linguistic information. 
Figure \ref{fig:sparse} 
shows the distributions of the two types of masks (weights and hidden activations, respectively) over the layers within BERT for for the semantics/syntax tasks.
We observe that the learned `semantic' mask zeros out fewer elements at higher layers in the network, while the `syntax' mask prefers to keep non-zero entries in lower layers. 
This suggests that semantic information may be captured mostly in higher layers of BERT, while syntactic information may be encoded in lower layers, consistent with observations in prior work \cite{tenney2019bert}.

\begin{figure}
\includegraphics[width=.4\textwidth]{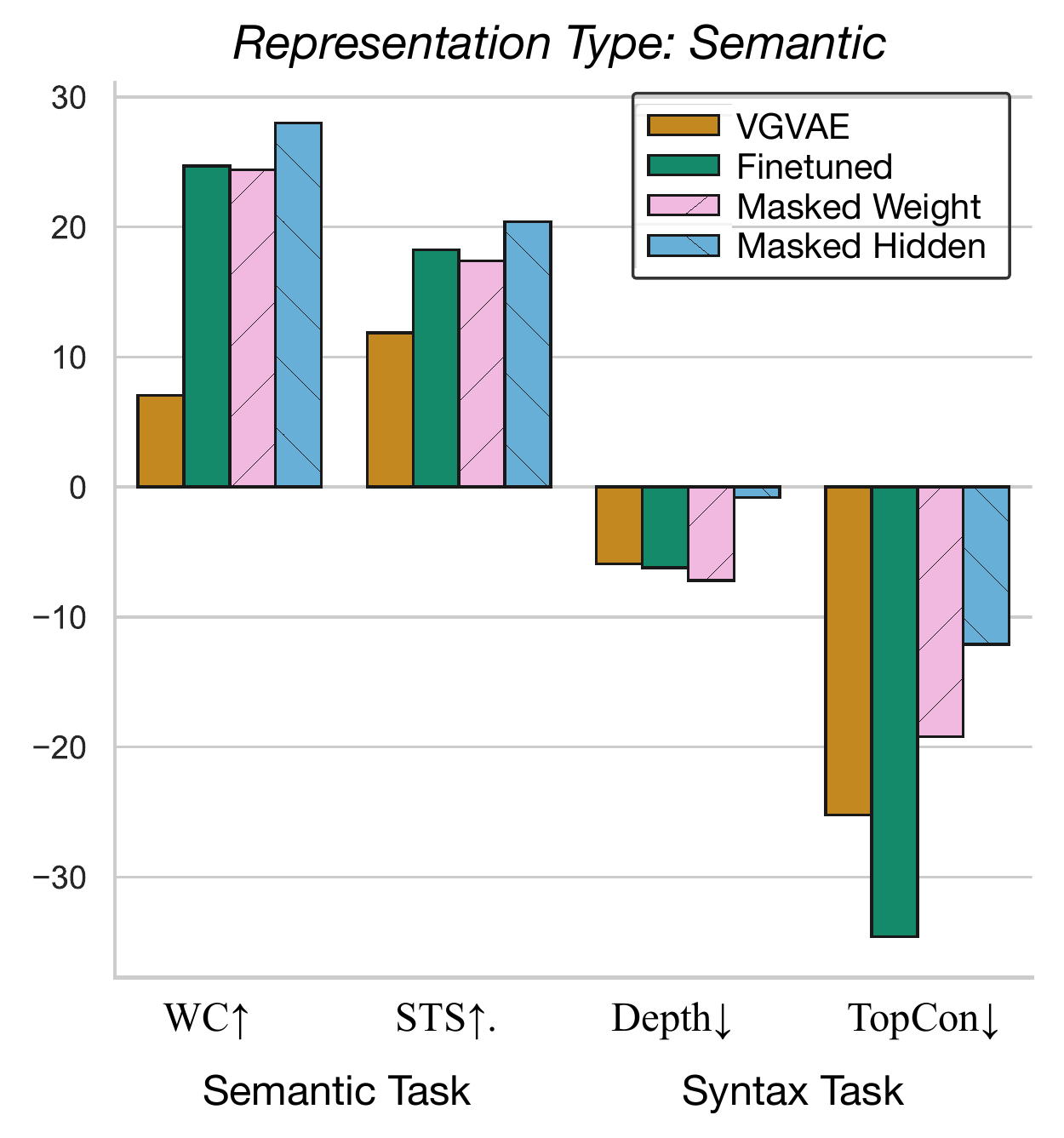}
\caption{Differences between the performances achieved via BERT embeddings and the disentangled model variants considered on semantics-oriented (WC, STS) and syntax-oriented (Depth, TopCon) tasks compared with BERT embeddings. We plot this difference with respect to the semantic embeddings induced by the models.}
\label{fig:rep-sem}
\end{figure}

\begin{figure*}
\begin{center}
\includegraphics[width=0.8\textwidth]{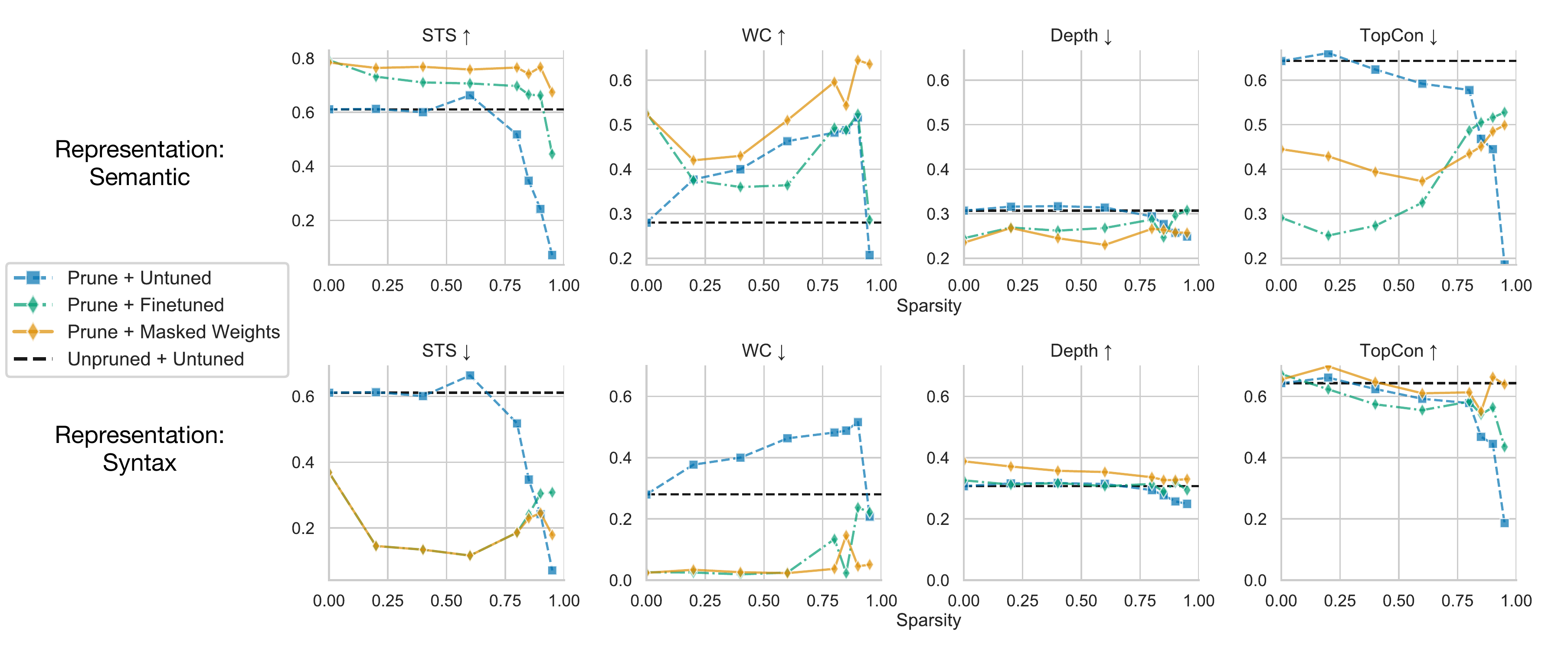}
\end{center}
\caption{Model performance as a function of the degree of pruning. The $x$-axis corresponds to the subnetwork sparsities (percent of weights dropped), while the $y$ axes are performance measures --- accuracy for all tasks except for STS, where we report Pearson's correlation. We compare the performance of models trained on the semantic (top) and syntax representations (bottom) learned by the disentangling strategies considered, after pruning to varying levels of sparsity.
}
\label{fig:prune}
\end{figure*}

\subsection{Semantic Representation Performance (vs. BERT)}
\label{sec:sem-rep}

We show the signed  differences  between the performance achieved on semantics- and syntax-oriented tasks by BERT embeddings (we mean-pool over token embeddings) and the ‘semantic’ representations from the disentangled models in figure \ref{fig:rep-sem}.

\subsection{Model performance with iterative magnitude pruning}
\label{sec:mag_pru}
We report full results of combining our method with magnitude pruning to uncover sparse sub-networks in Figure \ref{fig:prune}. 
We compare our method to several alternative pruning strategies: Standard magnitude tuning applied to BERT, without additional tuning (Pruned + Untuned), and a method in which after magnitude pruning we resume finetuning of the subnetwork for a fixed number of steps, using the aforementioned loss function (Pruned + Finetuned).

\begin{table}
\small
\centering
       \begin{tabular}{ lcc } 
       \toprule

        \textbf{Layers Masked} & \textbf{Sentiment} $\uparrow$ & \textbf{Leakage (Genre)} $\downarrow$  \\
        \midrule
        
     Last 3 & 86.3 & 73.5 \\ 
     Last 6 & 88.0 & 72.0   \\ 
     Last 9 & 88.0 & 72.8    \\ 
     All 12  & 87.5 & 72.3    \\ 
        \bottomrule
    \end{tabular}
    \caption{Performance of model (Masking Weights) when masking the last 3, 6, 9 or 12 layers of BERT.}
    \label{table:mask-layers}
\end{table}

\begin{table}
\small

\centering
       \begin{tabular}{ lcc } 
       \toprule

        \textbf{$\alpha$} & \textbf{Sentiment} $\uparrow$ & \textbf{Leakage (Genre)} $\downarrow$  \\
        \midrule
        
     0.5 & 83.8 & 74.8 \\ 
     1.0 & 86.3 & 73.0   \\ 
     2.0 & 88.0 & 72.0    \\ 
     5.0  & 87.8 & 70.3    \\ 
        \bottomrule
    \end{tabular}
        \caption{Performance of model (Masking Weights) when choosing different $\alpha$ for the triplet loss.}
    \label{table:alphas}
\end{table}

\subsection{Additional Experiments:Perturbation Study of Hyper-parameters}

We report model performance when masking different number of layers of BERT (Table \ref{table:mask-layers}) and when choosing different values for $\alpha$(Table \ref{table:alphas}).

\begin{table*}
\centering
\small
       \begin{tabular}{lccc} 
       \toprule
       & \textbf{Finetuned} & \textbf{Adversarial} & \textbf{Masking Weights} \\
       \midrule
\multicolumn{4}{l}{\textbf{Strong Correlation}: 15\% of "Drama" Reviews are Positive} \\
        \midrule
        
     Avg.(Sentiment) & 84.3 & 84.5 & 87.8 \\ 
     Worst (Sentiment)& 69.3 & 72.0 & 77.0 \\ 
        \midrule
    \multicolumn{4}{l}{\textbf{Moderate Correlation:} 25\% of "Drama" Reviews are Positive} \\        
        \midrule
     Avg.(Sentiment) & 86.8 & 86.5 & 86.3 \\ 
     Worst (Sentiment)& 72.0 & 76.8 & 80.5 \\ 
        \midrule
    \multicolumn{4}{l}{\textbf{No Correlation:} 50\% of "Drama" Reviews are Positive} \\
        \midrule
        
     Avg.(Sentiment) & 88.5 & 88.0 & 86.8 \\ 
     Worst (Sentiment)& 86.0 & 87.0 & 85.3 \\ 
        \bottomrule
    \end{tabular}
     \caption{Performance of our model (Masking Weights) compared to baseline models with varying degree of correlation in the training set.}
    \label{table:correlation}
\end{table*}

\subsection{Model performance with varying degree of correlation in the training set}

We report the comparison of our model (Masking Weights) with two baselines (Finetuned and Adversarially trained BERT) with varying degree of correlation in the training set. The task is sentiment classification and we control the correlation between sentiment and genre into 3 different settings: strong, moderate and weak (if any) correlation. We report the results in Table \ref{table:correlation}. Our model significantly outperforms the baselines when the correlation is strong, and the advantage begins to diminish as the correlation becomes weaker, as we would expect.

\end{document}